\pdfoutput=1

\documentclass[11pt]{article}

\usepackage{acl2023}

\usepackage{times}
\usepackage{latexsym}

\usepackage[T1]{fontenc}

\usepackage[utf8]{inputenc}

\usepackage{microtype}

\usepackage{mathtools}
\usepackage{subcaption}
\usepackage{framed}
\usepackage{amsfonts}
\usepackage{booktabs}
\usepackage{enumitem}

\usepackage{color, colortbl}
\definecolor{ao(english)}{rgb}{0.0, 0.5, 0.0}
\definecolor{bondiblue}{rgb}{0.0, 0.58, 0.71}

\newtheorem{character}{Character}

\begin{document}

\title{Exploring Automatically Perturbed Natural Language Explanations in Relation Extraction}

\author{Wanyun Cui \\
  Shanghai University of Finance and Economics \\
  \texttt{cui.wanyun@shufe.edu.cn} \\ \And
  Xingran Chen \\
  University of Michigan \\
  \texttt{chenxran@umich.edu} \\}

\maketitle

\begin{abstract}

Previous research has demonstrated that natural language explanations provide valuable inductive biases that guide models, thereby improving the generalization ability and data efficiency. In this paper, we undertake a systematic examination of the effectiveness of these explanations. Remarkably, we find that corrupted explanations with diminished inductive biases can achieve competitive or superior performance compared to the original explanations. Our findings furnish novel insights into the characteristics of natural language explanations in the following ways: (1) the impact of explanations varies across different training styles and datasets, with previously believed improvements primarily observed in frozen language models. (2) While previous research has attributed the effect of explanations solely to their inductive biases, our study shows that the effect persists even when the explanations are completely corrupted. We propose that the main effect is due to the provision of additional context space. (3) Utilizing the proposed automatic perturbed context, we were able to attain comparable results to annotated explanations, but with a significant increase in computational efficiency, 20-30 times faster.

\end{abstract}

\section{Introduction}
\label{sec:intro}

The application of neural networks in NLP has been a great success. However, the opaque nature of neural network mechanisms raises concerns regarding the reliability of their inferences and the potential for superficial pattern learning. This has led to severe issues with the generalization ability and vulnerability of neural networks.~\cite{jia2017adversarial,rajpurkar2018know}.

One common attempt to address the opacity of neural networks is to guide them with explanations. Researchers propose that by connecting the model and the inductive bias from explanations, the reliability of neural network inferences can be improved. This approach has been successfully applied in relation extraction tasks, as demonstrated in previous studies such as ~\cite{srivastava2017joint,hancock2018training,murty2020expbert}.


\begin{figure*}[!tb]
\begin{subfigure}[b]{0.53\textwidth}
\centering
    \begin{framed}
\small
\vspace{-0.25cm}
\begin{align*}
& \text{x: {\it Robert} and {\it Julie} had a terrible honeymoon last month.} \\
& \text{y: Spouse} \\
&\text{{\bf Explanation}:} \text{ $o_1$ and $o_2$ went on a \textcolor{bondiblue}{\bf honeymoon}.} \\
\midrule 
&\text{{\bf Corrupted explanation}:} \text{ $o_1$ and $o_2$ went on a \textcolor{ao(english)}{\bf frog}.} \\
&\text{{\bf Perturbed context}:} \text{ $\mathrm{o_1\:\: [M]_1\:\: [M]_2\:\: \cdots\:\: [M]_m \:\: o_2}$}
\end{align*}
\vspace{-0.8cm}
\end{framed}
\caption{Guide model with explanations.}
\label{fig:corrupt_example}
\end{subfigure}
\begin{subfigure}[b]{0.225\textwidth}
	\centering
		\includegraphics[scale=.35]{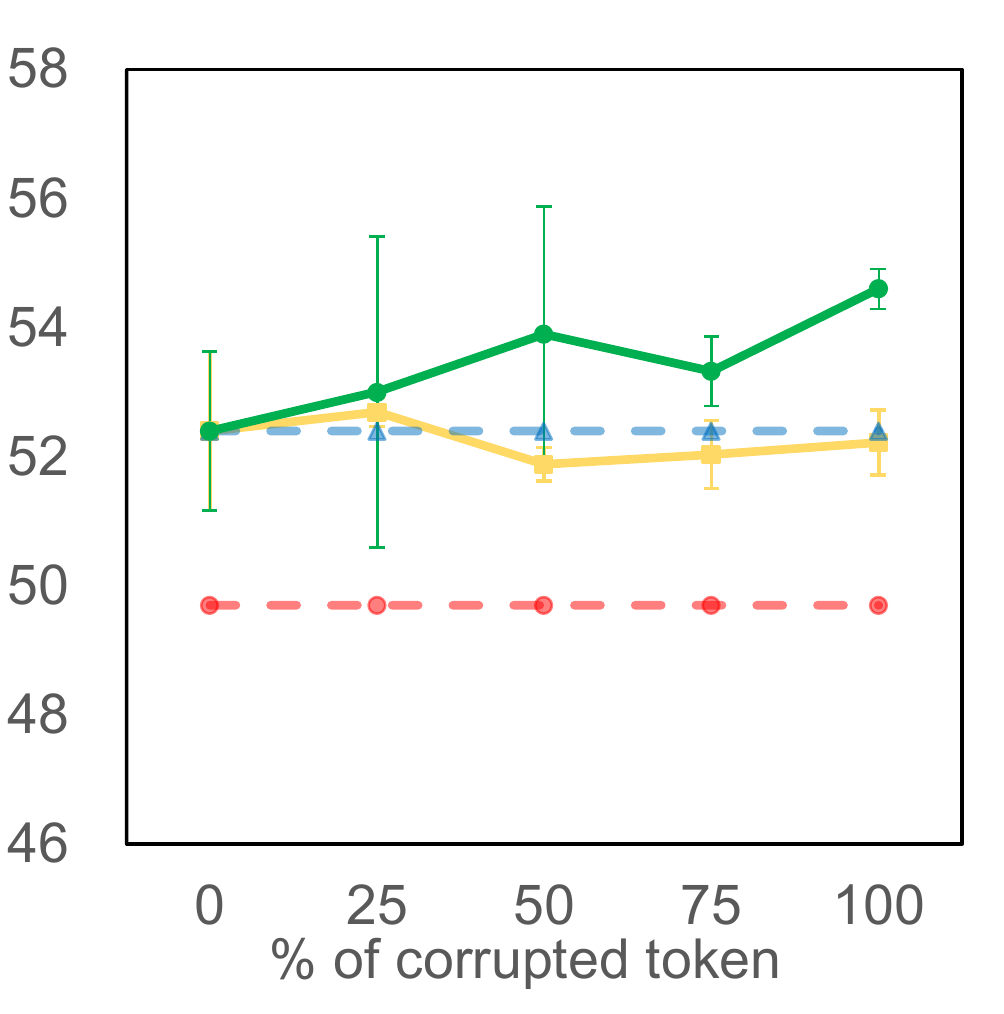}
\caption{Disease} 
\label{fig:random:freezed:disease}
\end{subfigure}
\begin{subfigure}[b]{0.225\textwidth}
	\centering
		\includegraphics[scale=.35]{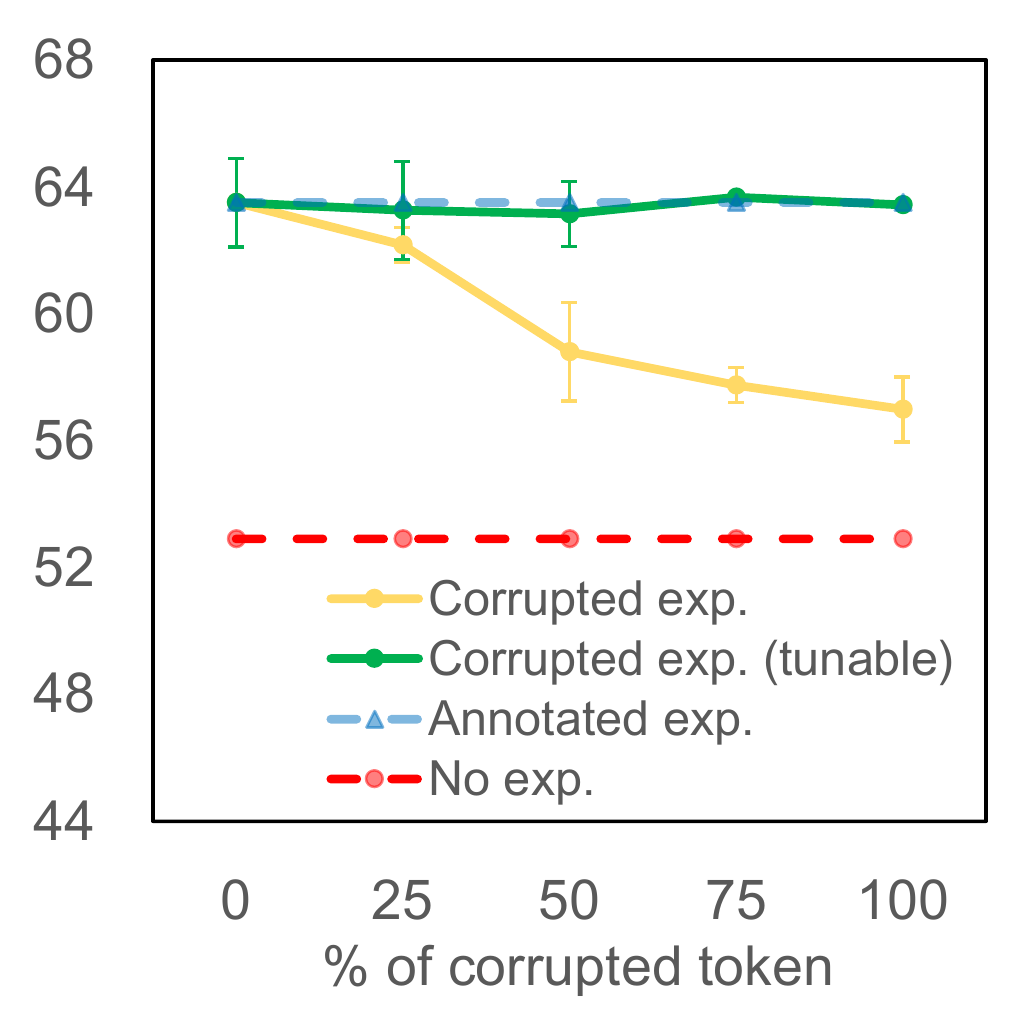}
\caption{Spouse} 
\label{fig:random:freezed:spouse}		
\end{subfigure}
    \caption{Corrupted explanations and their impact under frozen LM setting. Our results show that the random corruption does not result in a decrease in performance on the Disease dataset. While corruptions do lead to a reduction in the effect on the Spouse dataset, we find that even when explanations are 100\% corrupted, they still result in an improvement over the baseline without explanations. These findings are an average of 3 runs.}
	\label{fig:random:freezed}
\end{figure*}

Earlier approaches relied on explanations from semantic parsers~\cite{srivastava2017joint,hancock2018training}, which incurs a high annotation cost. The recently proposed approach, ExpBERT~\cite{murty2020expbert}, was a breakthrough in its ability to directly incorporate {\it natural language explanations}. For example, in Fig.~\ref{fig:corrupt_example}, {\it $o_1$ and $o_2$ went on a honeymoon} can be used as one explanation to guide the recognition of the spousal relation. ExpBERT with annotated explanations achieves $63.5\%$ accuracy in the spousal relation extraction dataset, while BERT without explanations only achieves $52.9\%$.

Considering the simple mechanism of ExpBERT, such improvement is quite surprising. ExpBERT simply concatenates explanations  with the original text before being encoded by a language model. Based on the success of ExpBERT, one might conclude that text concatenation and pre-trained language models are sufficient for integrating the inductive bias from natural language explanations and guide models to make a sound inference. On the other hand, as exemplified by the history of deep learning, introducing extra inductive biases into neural networks is never trivial. Due to the strong generalization ability of neural networks, introducing inductive biases by humans is often surpassed by simpler models or even random inductive biases~\cite{xie2019exploring,touvron2021resmlp,tay2021synthesizer}. 

In this paper, starting from investigating the working mechanism of ExpBERT, we study how explanations guide and enhance the model effect. 
We first propose a simple strategy to control the inductive bias of explanations by the lens of {\it corrupted explanations}, wherein some words of annotated explanations are replaced by random words. We show an example of the corrupted explanation in Fig.~\ref{fig:corrupt_example}, where the word {\it honeymoon} is replaced by the random word {\it frog}. Obviously, the explanations will provide less valid information after random corruption.

We show the effect of corrupted explanations in Fig.~\ref{fig:random:freezed:disease} and Fig.~\ref{fig:random:freezed:spouse}. On both datasets, adding explanations shows a clear improvement to the baseline with no explanation. Surprisingly, however, reducing the inductive bias of explanations has almost no effect on the improvement of explanations on the Disease dataset. On the Spouse dataset, although the improvement decreases as the corruption increases, the effect of 100\% corrupted explanations is still better than no explanations. These results suggest that the effect of explanations should not be entirely attributed to their inductive bias.

With comprehensive experiments of corrupted explanations in \S\ref{sec:analysis}, we identified the following characters of natural language explanations:
\begin{itemize}[leftmargin=*]
    \item{\bf Sensitivity} The effect of natural language explanations is sensitive to the training style and the downstream dataset. The
    previously observed improvement in accuracy and data efficiency~\cite{murty2020expbert} is only applicable for frozen language models. For fine-tunable language models, the improvement becomes less significant and does not generalize to all datasets.
    \item{\bf Cause} The effect of the natural language explanations comes from the extra context space, rather than their inductive bias. Given enough context space, there is no significant effect decrease over annotated explanations even if they are completely corrupted.
    \item{\bf Parameter search helps} The manual annotations provide a good initialization for that context - although it can also be obtained via parameter search. If we randomly initialize the extra context and fine-tune it with downstream datasets, it achieves competitive or superior results over annotated explanations.
\end{itemize}

The above findings motivate us to further investigate and improve natural language explanations. To get rid of the potential entanglement of the existing vocabulary, we further proposed the {\it perturbed context} as the substitute, which only contains randomly initialized embeddings. We conducted different variants of perturbed contexts to investigate how natural language explanations work. We find that full-rank random contexts achieve competitive results with annotated explanations but are 18-29 times faster.

\section{Background and Experimental Setup}
\subsection{Problem Definition}
We consider the relation extraction task, which is frequently used for explanation guidance evaluation. Given $x=(s,o_1,o_2)$, where $s$ is the target sentence, $o_1$ and $o_2$ are two entities which are substrings of $s$, our goal is to predict the relation $y$ between $o_1$ and $o_2$.

Additionally, a set of natural language explanations $\mathcal{E}=\{e_1,\cdots,e_n\}$ are annotated to capture relevant inductive bias for this task. This setting follows~\cite{murty2020expbert}. Note that these explanations are designed to capture the global information for all samples in this task, rather than for each example.

For example, for the spousal relationship, {\it ``$o_1$ and $o_2$ went on a honeymoon''} is a valid explanation used in ExpBERT. We claim that this explanation constitutes a global inductive bias, and whether $o_1$ and $o_2$ went on a honeymoon will be seen as a feature to determine their spousal relationship for all samples. Similar global feature settings are also used in previous studies~\cite{srivastava2017joint,hancock2018training,murty2020expbert}.

\subsection{Guiding Language Models with Explanations}

{\bf Introducing annotated explanations} ExpBERT~\cite{murty2020expbert} is a state-of-the-art model for introducing annotated natural language explanations for relation extraction. In representing the samples $x=(s,o_1,o_2)$, ExpBERT first splices $s$ with each natural language interpretation $e_i$ by a separator $[SEP]$. Then it represents this splice by BERT:
\begin{equation}
\small
    \mathcal{I}(s,e_i) = \mathrm{BERT( [CLS],s,[SEP],e_i, [SEP])} \in \mathbb{R}^d
\end{equation}
where $d$ represents the dimension of the hidden states of BERT. ExpBERT concatenates the $n$ representations (i.e. $\mathcal{I}(s,e_1)$ $\cdots \mathcal{I}(s,e_n)$ ) as the explanation augmented representation of $s$. The representation is then classified to the corresponding relation by a MLP classifier.

{\bf Introducing corrupted explanations} work similarly to ExpBERT, except that a certain fraction of tokens in explanations are replaced by random tokens. The more tokens to be corrupted, the less inductive bias the explanation retains.

{\bf No explanation} We also compare with the vanilla language model without introducing explanations. We use BERT as the language model by default.


\begin{figure*}[htb]
\centering
\begin{subfigure}[b]{0.3\textwidth}
	\centering
		\includegraphics[scale=.35]{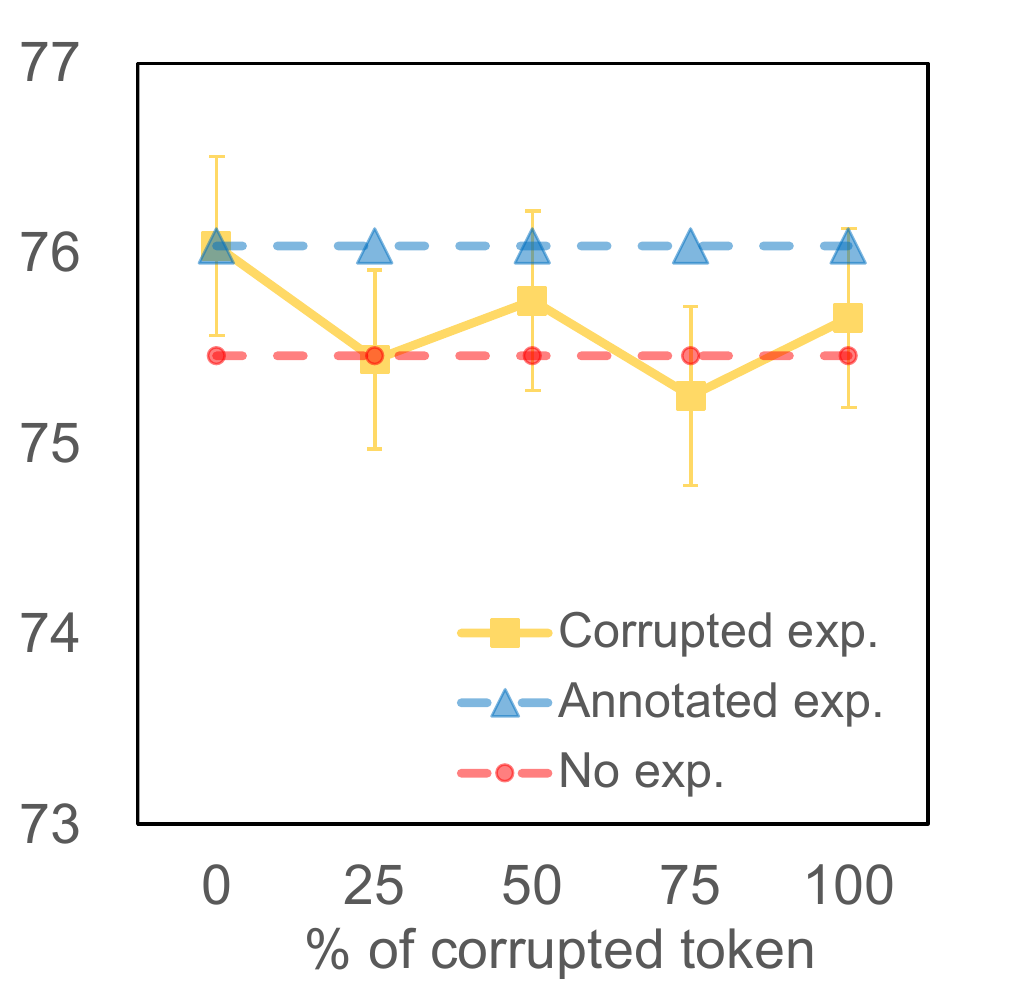}
\caption{Spouse.}
\label{fig:random:spouse}
\end{subfigure}
\hspace{0.2cm}
\begin{subfigure}[b]{0.3\textwidth}
	\centering
		\includegraphics[scale=.35]{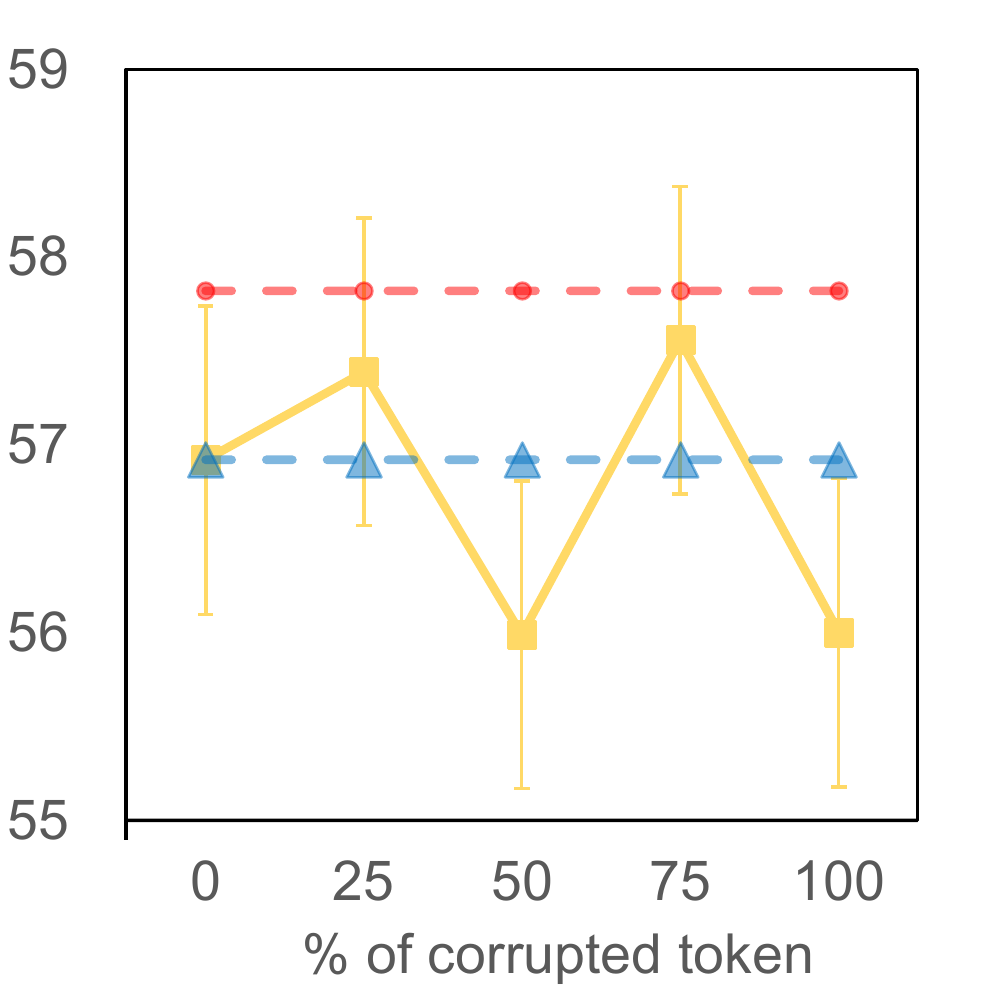}
\caption{Disease.}
\label{fig:random:disease}
\end{subfigure}
\begin{subfigure}[b]{0.3\textwidth}
	\centering
		\includegraphics[scale=.35]{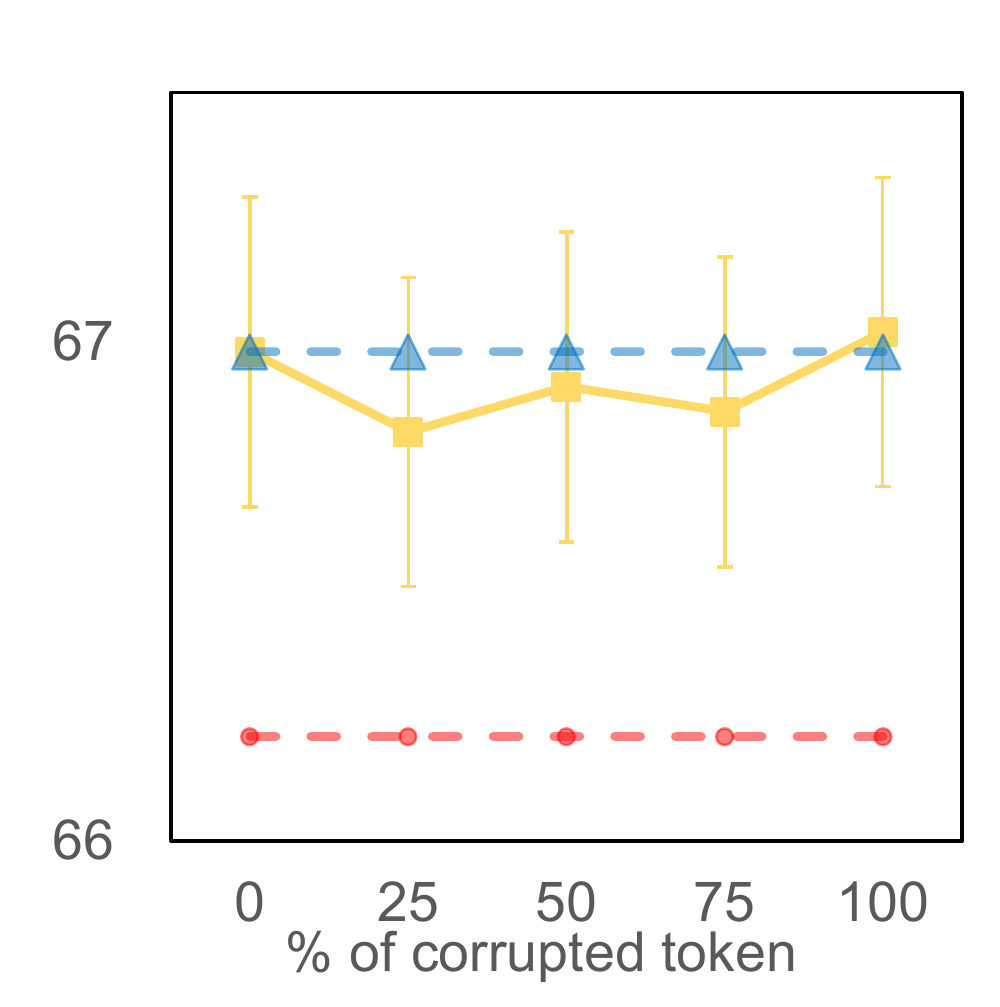}
\caption{TACRED.}
\end{subfigure}
	\caption{Results for fine-tunable language models. Annotated explanations show unstable and minor improvement over vanilla language models.
	The effectiveness did not significantly decrease after random corruptions. 
	}
	\label{fig:random}
\end{figure*}

\begin{figure*}[!htb]
 \centering
 \begin{subfigure}[b]{0.3\textwidth}
 	\centering
 		\includegraphics[scale=.35]{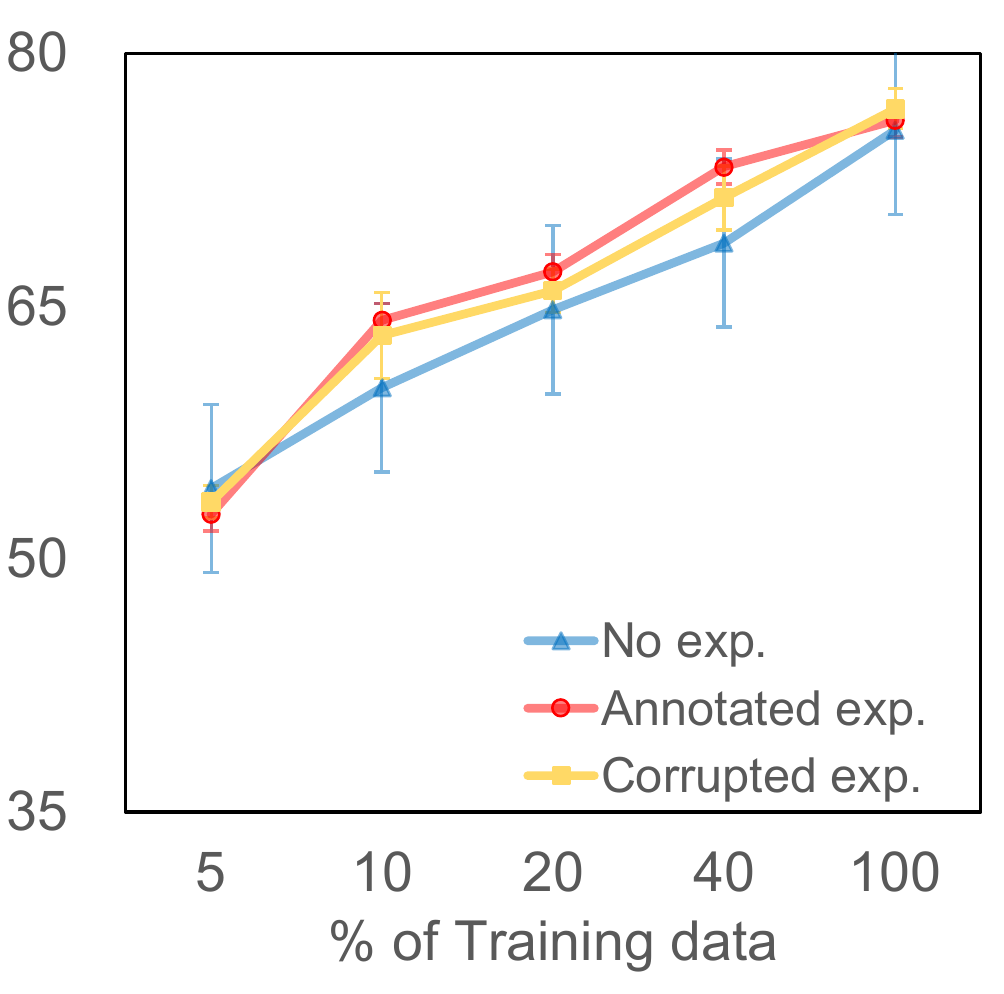}
 \caption{Spouse.}
 \end{subfigure}
 \begin{subfigure}[b]{0.3\textwidth}
 	\centering
 		\includegraphics[scale=.35]{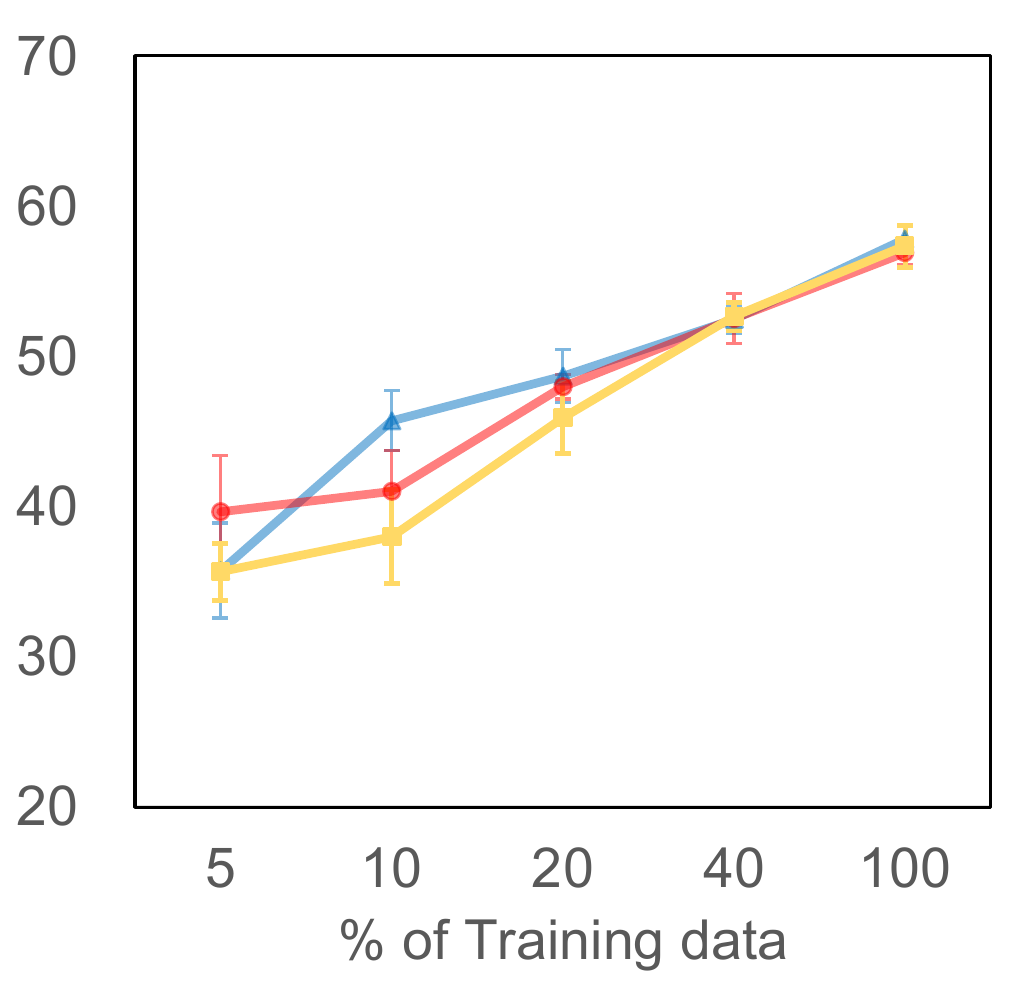}
 \caption{Disease.}
 \end{subfigure}
 \begin{subfigure}[b]{0.3\textwidth}
 	\centering
 		\includegraphics[scale=.35]{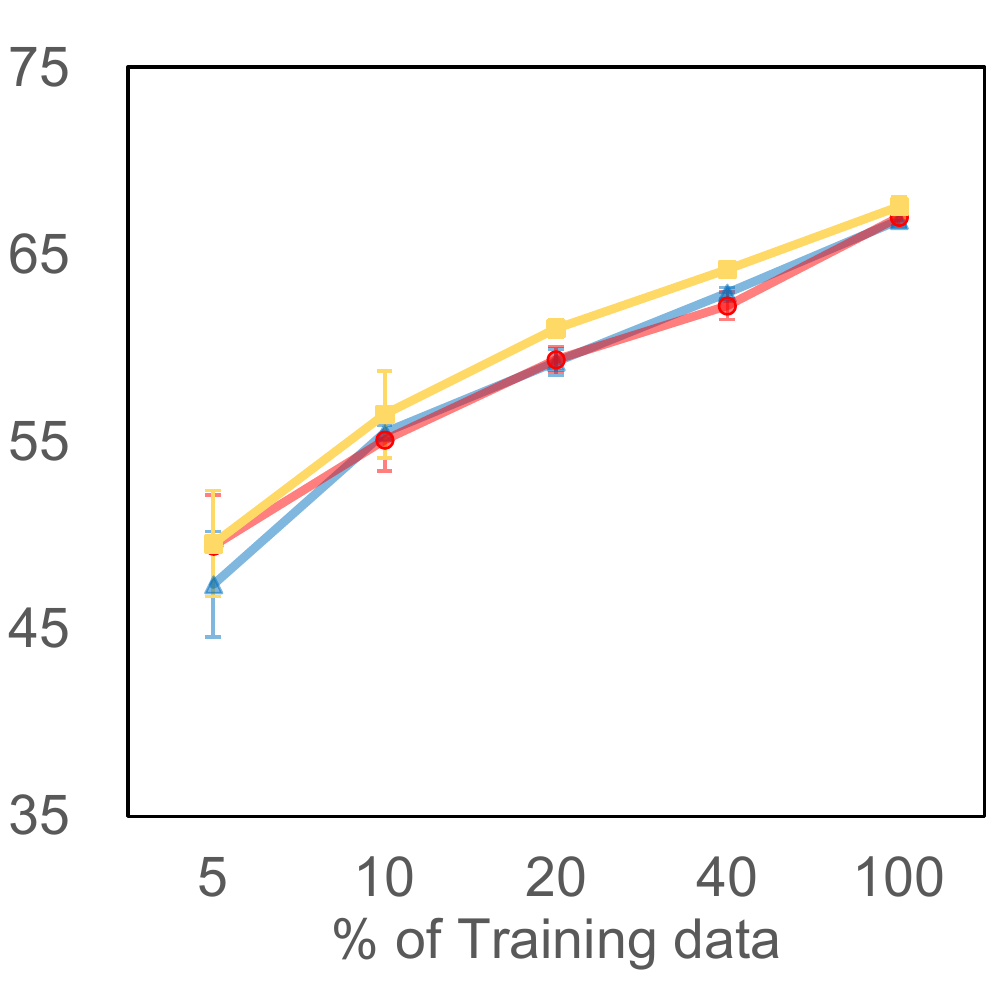}
 \caption{TACRED.}
 \end{subfigure}
 	\caption{For fine-tunable language models, introducing explanations does not improve data efficiency. 
  }
 	\label{fig:dataefficiency}
 \end{figure*}

\subsection{Training Styles}

We consider two training styles: fine-tuning and frozen language models. In {\bf fine-tuning}, all parameters will be dynamically updated through back-propagation. In {\bf frozen language models}, all parameters in the language model BERT are frozen after being pre-trained by an additional corpus (i.e., MultiNLI~\cite{N18-1101}), allowing only tuning the MLP classifier. This setting is used in~\cite{murty2020expbert}.

\subsection{Datasets}

\begin{table}[!tb]
\small
\centering
\caption{Dataset statistics.}
\begin{tabular}{ccccc} \toprule
Dataset & Train & Val   & Test & \#Exp.  \\ \hline
Spouse & 22055 & 2784  & 2680 & 41  \\
Disease & 6667  & 773   & 4101 & 29 \\
TACRED  & 68124 & 22631 & 15509 & 128 \\ \bottomrule
\end{tabular}
\label{tab:stat}
\end{table}

We follow~\cite{murty2020expbert} to use three benchmarks: Spouse~\cite{hancock2018training}, Disease~\cite{hancock2018training}, and TACRED~\cite{zhang2017position}. 
We use the annotated natural language explanations provided by~\cite{murty2020expbert} for the baselines, except TACRED, whose natural language explanations are not published.
Therefore, we manually annotated 128 explanations as in~\cite{murty2020expbert}. The statistics of the datasets is shown in Table~\ref{tab:stat}. More details of the implementations are demonstrated in Appendix~\ref{sec:hyper}.

\section{Characterizing Natural Language Explanations}
\label{sec:analysis}

In this section, we make a thorough experimental study of the characters of natural language explanations. We plot the accuracy of different settings in Fig.~\ref{fig:random:freezed} (frozen) and Fig.~\ref{fig:random} (fine-tunable).

\begin{character}
The effect and data efficiency of natural language explanations are sensitive to the training style and the dataset. 
\end{character}

{\bf Effect improves for frozen language models.} For frozen language models (Fig.~\ref{fig:random:freezed}), introducing annotated natural language explanations significantly improves the effect over models without explanations. This is in line with the finding in~\cite{murty2020expbert}. 

{\bf The improvement becomes less significant for fine-tuning language models and varies across datasets.} However, in the more common setting where all parameters are fine-tuned, the effect of annotated explanations is unstable (Fig.~\ref{fig:random}). On the Spouse and TACRED datasets, introducing annotated explanations has accuracy improvement, but not on Disease. The improvement is less significant compared to the frozen language models. This challenges the perception in previous work that introducing natural language explanations has significant effects~\cite{murty2020expbert}.

{\bf Data efficiency only holds for frozen language models} As to data efficiency, previous study~\cite{murty2020expbert} found that the effect of a model trained on a full amount of data without explanations can be achieved by using a small subset (e.g. $5\%$) of training data with explanations. However, they only verified this in frozen language models. We analyzed the data efficiency in fine-tunable language models. We vary the proportion of training data on different datasets. The results are shown in Fig.~\ref{fig:dataefficiency}. 
Similar to the accuracy experiments, we find that the data efficiency of natural language explanations does not generalize to fine-tunable language models. The natural language explanation does not improve the data efficiency, nor does the corrupted explanation.

\begin{character}
Reducing the inductive bias of annotated explanations does not significantly decrease the effect.
\end{character}
We also demonstrate the results of corrupting a certain fraction of tokens in annotated explanations in Fig.~\ref{fig:random:freezed} and Fig.~\ref{fig:random}. The results are surprising: after corrupting with random words, the performance does not drop in most cases. Even when we replace 100\% of the words in explanations with random words, the results are still competitive with the original explanations. This phenomenon was observed in all three datasets in the fine-tuning setting. Obviously, the 100\% corrupted explanations do not provide any valid inductive bias for the model.

Interestingly, for the frozen language models, the results for Spouse and Disease diverge. In Fig.~\ref{fig:random:freezed:disease}, corrupting the explanations does not reduce the effect of Disease. The results of Spouse in Fig.~\ref{fig:random:freezed:spouse}, however, shows that randomly corrupting the explanations does reduce the effect.  Fig.~\ref{fig:random:freezed:spouse} conflicts with other settings. We further investigate this exception below.

\begin{character}
    Parameter search over the corrupted tokens makes corrupted explanations comparable with annotated explanations in frozen language models.
\end{character}

In the frozen language models, both the language model and the corrupted explanations are not fine-tunable. The randomly corrupted explanations may not be well-initialized if no optimization is allowed. We expect that the corrupted explanation will still be effective after proper initialization.

To verify this, we slightly modified the training strategy to allow the word embeddings of the corrupted words to be fine-tuned, while still freezing other parameters of the language model. That is, we search for the parameters for the corrupted explanation. Note that, during this process, no annotated explanation is used.

We show the results with tunable corrupted tokens in Fig.~\ref{fig:random:freezed}. We find that while not introducing extra annotations, the parameter search for the corrupted explanation performs competitively (on Spouse) or surpasses (on Disease) the annotated explanations. The results suggest that random initialization does not necessarily perform well. On the other hand, manually annotated explanations serve as a good initialization for the extra context.


\section{How Do Explanations Work? An Investigation via Perturbed Context}
\label{sec:how}

The effect of corrupted explanations and natural language explanations reminds us to investigate how explanations work through their commonality: they both provide extra contexts. We hypothesize that the real factor at play is the context space provided by (corrupted) explanations, rather than the inductive bias. 
Hence, in this section, we systematically analyze how explanations work regarding the extra context space. To address this, we present perturbed context. 

\subsection{Definition of Perturbed Context}
\label{sec:method:definition}
Our approach is inspired by the experiments of randomly corrupted explanations in \S~\ref{sec:analysis}. According to the experiments, inductive biases of the explanations are not important, but rather we need a fine-tunable context to enrich the text representation. Therefore, we propose the perturbed context without any annotation to provide a fine-tunable context. We define a perturbed context $e_v$ in the following form:
\begin{equation}
\small
    \mathrm{e_v \coloneqq o_1\:\: [M]_1\:\: [M]_2\:\: \cdots\:\: [M]_m \:\: o_2}
\end{equation}
where $o_1$ and $o_2$ are placeholders for the two entities. We denote the embedding of $[\mathrm{M}]_i$ as $emb([\mathrm{M}]_i)$. We use $[\mathrm{M}]_i$ as the normal token input in language models. We set $m=4$ in this paper. $emb([\mathrm{M}]_i)$ is randomly initialized without semantics.  

\subsection{Variants of Perturbed Contexts}
\label{sec:method:variants}
As the extra context is the commonality between annotated explanations and corrupted explanations, we investigate how explanations work w.r.t. extra contexts. We introduce several variants of the perturbed contexts. We are particularly interested in (1) whether the perturbed context should be conditioned on the input $x$; (2) how the flexibility of the context space affects the results. The flexibility is controlled via factorization. We refer to the implementation of Synthesizer~\cite{tay2021synthesizer}, which is a recent study that revisits the inductive biases in attention.

{\bf Randomly perturbed contexts} We consider the simplest form of the perturbed context, which consists of independent perturbed tokens that are randomly initialized. We set the perturbed contexts to be global and task-specific, rather than  sample-specific. The randomly perturbed context is not conditioned on any input tokens. 

Let $\mathrm{M \in \mathbb{R}^{m \times d}}$ be a randomly initialized matrix. The embeddings of tokens in the randomly perturbed context are defined as:
\begin{equation}
\small
   \mathrm{ emb_{rand}([M]_i) = M_i}
\end{equation}
The randomly perturbed context has $\mathrm{m \times d}$ parameters. These parameters can either be trainable or kept fixed (denoted as {\bf fixed random}).

{\bf Conditional perturbed context} We also consider constructing perturbed contexts that are conditioned on each sample $x$. Here we adopt $F_i()$, a parameterized function, for projecting $x$ to the embedding space of the perturbed tokens.
\begin{equation}
\small
    \mathrm{emb_{cond}([M]_i) = F_i(x_{pool})}
\end{equation}
where $\mathrm{x_{pool} \in \mathbb{R}^d}$ is the pooled BERT output for $x$. In practice, we use the multi-layer perceptron as $F(\cdot)$.

{\bf Factorized models} We investigate the effect of the flexibility of the context. We refer to the factorization method in ALBERT~\cite{lan2019albert}. We map the original embedding of the perturbed token to a lower dimensional space, and then project it back to the original embedding space. The size of the intermediate space reflects flexibility. Following this idea, we further design the following variants.

{\bf Factorized randomly perturbed context} We factorize the embedding of the randomly perturbed context by:
\begin{equation}
\small
    \mathrm{emb_{f\_rand}([M]_i) = W_{fr} MF_i}
\end{equation}
where $\mathrm{MF \in \mathbb{R}^{m \times l}}$ is a lower dimensional embedding matrix,
$\mathrm{W_{fr} \in \mathbb{R}^{d\times l}}$. We first use $MF_i$ to represent the lower dimensional space of size $l (l<d)$, then we project it back to the normal embedding space of size $d$. 

{\bf Factorized conditional perturbed context} Similarly, we also factorize the embedding of the conditional perturbed context by:
\begin{equation}
\small
    \mathrm{emb_{f\_cond}([M]_i) = W_{fc2} (W_{fc1} F_i(x_{pool}))}
\end{equation}
where $\mathrm{W_{fc1} \in \mathbb{R}^{l\times d}}$,$\mathrm{W_{fc2} \in \mathbb{R}^{d\times l}}$. 

{\bf Mixture explanations} We consider combining the perturbed context and manually annotated explanations to see if the two kinds of contexts complement each other. To ensemble the explanations, we add the perturbed context to the annotated explanation list. 

{\bf Exploiting the perturbed context}
We use the perturbed context as the context of the original sentence. Specifically, we use the pre-trained language model BERT to represent the target sample $x=(s,o_1,o_2)$. We construct $e_v$ as a context for $s$ by replacing the placeholders with $o_1$ and $o_2$, respectively. Then we use the $\mathrm{[SEP]}$ token as a separator to combine $s$ and $e_v$. In this way, we convert the representation of $s$ into the representation of $s+e_v$. We use the default setting in BERT to represent the sentence pair, i.e., using the final output of the $\mathrm{[CLS]}$ token as the representation of sentence pairs:
\begin{equation}
\small
    \mathcal{I}(s,e_v) = \mathrm{BERT( [CLS],s,[SEP],e_v, [SEP])}
\label{eqn:input}
\end{equation}

ExpBERT uses $n$ explanations to improve the model, while we found one perturbed context achieves near-optimal results. We will give more experimental evidence in \S~\ref{sec:main_result} and Appendix~\ref{sec:exp:multi}.

\begin{table*}[!tb]
\small
\centering
\begin{tabular}{l c cc@{\hskip 0.3in}cc@{\hskip 0.3in}ccc} 
\toprule
Model & Annotated  & \multicolumn{2}{c}{\textbf{Spouse}} & \multicolumn{2}{c}{\textbf{Disease}}  & \multicolumn{2}{c}{\textbf{TACRED}} & \textbf{Avg.} $\textbf{F1}$ \\
 & & $\textbf{F1}$ & \textbf{Time} & $\textbf{F1}$ & \textbf{Time} & $\textbf{F1}$ & \textbf{Time} \\ \hline
BabbleLabble & Yes      & 50.1 $\pm$ 0.00 & - & 42.3 $\pm$ 0.00 & - & - & - & - \\
NeXT  & Yes   & - & - & - & - & 45.6 & - & - \\ \midrule
BERT  & No                  & 75.5 $\pm$ 0.59 & $1.0\times$ & \textbf{57.8 $\pm$ 0.90}  & $1.0\times$ &  66.8 $\pm$ 0.31     & $1.0\times$  & 66.7 \\
ExpBERT \dag    & Yes       & 76.0 $\pm$ 0.47 & $28.5\times$ & 56.9 $\pm$ 0.82 & $20.1\times$ & 67.0 $\pm$ 0.14 & $32.1\times$ & 66.6 \\
\textbf{PC (BERT)}  & & & & & & & \\
\quad + \textbf{R} & No              & 76.7 $\pm$ 1.50 & $1.1\times$ & 57.3 $\pm$ 1.57 & $1.1\times$ & \textbf{67.6 $\pm$ 0.55} & $1.1\times$ & \textbf{67.2} \\ 
\quad + \textbf{Fixed R}  & No        & 74.9 $\pm$ 0.81 & $1.1\times$ & 56.2 $\pm$ 0.75 & $1.1\times$ & 66.9 $\pm$ 0.75 & $1.1\times$ & 66.0 \\
\quad + \textbf{C}  & No              & 75.8 $\pm$ 1.13 & $1.1\times$ & 57.3 $\pm$ 0.50 & $1.1\times$ & 66.6 $\pm$ 0.41 & $1.1\times$ & 66.6 \\
\quad + \textbf{FR}  & No             & \textbf{77.1 $\pm$ 1.48} & $1.1\times$ & 57.0 $\pm$ 0.51 & $1.1\times$ & 67.3 $\pm$ 0.25 & $1.1\times$ & 67.1 \\
\quad + \textbf{FC}  & No             & 75.8 $\pm$ 0.16 & $1.1\times$ & 56.6 $\pm$ 0.61 & $1.1\times$ & 66.8 $\pm$ 0.44 & $1.1\times$ & 66.4 \\
Mixture  & Yes       & 76.3 $\pm$ 1.06 & $28.5\times$ & 56.5 $\pm$ 1.22 & $20.1\times$ & 66.5 $\pm$ 0.69 & $32.1\times$ & 66.4 \\ \midrule
RoBERTa  & No               & 77.1 $\pm$ 1.21 & $1.0\times$ & 58.8 $\pm$ 0.77  & $1.0\times$ & 69.2 $\pm$ 0.48   & $1.0\times$ & 68.4 \\
ExpRoBERTa \dag & Yes       & 75.9 $\pm$ 0.77 & $21.2\times$ & 57.0 $\pm$ 1.22 & $21.3\times$ & 68.6 $\pm$ 0.69 & $20.6\times$ & 67.2 \\
\textbf{PC (RoBERTa)} & & & & & & & \\
\quad + \textbf{R} & No           & 77.6 $\pm$ 0.52 & $1.1\times$ & 59.0 $\pm$ 0.55 & $1.1\times$ & \textbf{70.1 $\pm$ 0.37} & $1.1\times$ & 68.9 \\ 
\quad + \textbf{Fixed R}   & No    & \textbf{78.2 $\pm$ 0.65} & $1.2\times$ & \textbf{59.4 $\pm$ 0.86} & $1.1\times$ & 69.7 $\pm$ 0.43 & $1.0\times$ & \textbf{69.1} \\
\quad + \textbf{C}  & No           & 77.9 $\pm$ 0.56 & $1.3\times$ & 58.5 $\pm$ 0.50 & $1.2\times$ & 69.1 $\pm$ 0.24 & $1.2\times$ & 68.5 \\
\quad + \textbf{FR}  & No          & 77.5 $\pm$ 0.40 & $1.2\times$ & 57.0 $\pm$ 2.06 & $1.1\times$ & 69.3 $\pm$ 0.53 & $1.1\times$ & 68.0 \\
\quad + \textbf{FC}  & No          & 77.7 $\pm$ 1.01 & $1.3\times$ & 56.5 $\pm$ 1.12 & $1.2\times$ & 69.5 $\pm$ 0.15 & $1.2\times$ & 67.9 \\
Mixture  & Yes    & 75.9 $\pm$ 0.51 & $21.5\times$ & 57.1 $\pm$ 1.08 & $22.1\times$ & 68.7 $\pm$ 0.57 & $22.1\times$ & 67.2 \\
\bottomrule
\end{tabular}
\caption{Results for fine-tunable language models. Without introducing external annotated explanations, the perturbed context achieves competitive results with models with explanations. The efficiency of the perturbed context is significantly higher than its competitors. We denote the perturbed context as {\bf PC}. \textbf{R} means fine-tunable random. \textbf{C} means conditional. \textbf{Fixed R} means fixed random. \textbf{FR} means factorized random. \textbf{FC} means factorized conditional. Results are averaged over 5 runs. \dag: We fine-tune all parameters within ExpBERT and ExpRoBERTa.}
\label{tab:main}
\end{table*}



\subsection{Effect on Fine-Tunable Language Models}
\label{sec:main_result}
We show the effect of different approaches in Table~\ref{tab:main}. We also compare with the following baselines that also introduce explanations in text understanding: BabbleLabble~\cite{hancock2018training}, NeXT~\cite{wang2019learning}. 
Since ExpBERT is based on BERT, to make the comparison fair, we also use BERT as the language model by default. In addition, we also conducted experiments on RoBERTa~\cite{liu2019roberta}.



{\bf Effect of perturbed contexts} The proposed perturbed contexts overall achieve competitive performance with annotated explanations. This further verifies the claim in \S~\ref{sec:analysis}, that the natural language explanations mainly provide extra context, rather than the specific inductive bias. 

Among different variants of perturbed contexts, the fine-tunable random variant achieves the highest accuracy and outperforms the annotated explanations (ExpBERT) by $+1.15\%$ on average. We think this is because the annotated explanations have limited expressiveness and are cognitively biased. Thus a tunable context works better. 

{\bf Fine-tunable vs fixed} Surprisingly, we found that the fixed random variant performs competitively with the fine-tunable one on RoBERTa, and even achieves the highest accuracy in some settings. We think this is because the tunable language models provide sufficient flexibility to complement the flexibility of fixed perturbed contexts.

{\bf Global vs conditioned}
We found that the sample-specific variants (conditional and factorized conditional) have slight performance degradation compared to the global variant. We think this is because the explanation learned from the training data is prone to contain certain biases. This makes it hard to learn a sample-specific perturbed context generator with high generalization ability. In contrast, learning generalized global perturbed contexts is easier.
In terms of flexibility, conditional perturbed tokens are actually a projection of the representation of the original sentence $x$. This limits its flexibility.
This also corroborates our analysis in Appendix~\ref{sec:method:rationale} that we need a richer context to enhance the representation. 

{\bf Effect of mixture explanations} We evaluate the effectiveness of adding the randomly perturbed context to the annotated explanation list. The results in Table~\ref{tab:main} show that mixture explanations have a slight performance degradation over the randomly perturbed context. This indicates that manually annotated explanations do not complement with the randomly perturbed context.

{\bf Efficiency} Since the traditional approaches require the encoding of $n$ explanations, they have substantial extra training/inference time compared to the vanilla language models. For example, ExpBERT encodes $n$ explanations for each target sentence, which is extremely expensive when $n$ is large. (e.g. TACRED has 128 explanations). Our proposed perturbed contexts, on the other hand, only need to encode a single sentence consisting of the target sentence and the perturbed context. It is almost as efficient as encoding the original sentence. Therefore, the efficiency of our approach is substantially improved compared to the previous work. We show the average training time of different approaches in Table~\ref{tab:main}. The training time of our approach is almost the same as that of language models without explanations. While having competitive effects, our approach is about $20-30$ times faster than ExpBERT which introduces explanations.

\begin{table}[!tb]
\small
\centering
\begin{tabular}{lccc} \toprule
Model      & \textbf{Spouse} & \textbf{Disease}  \\ \hline
BabbleLabel-LangExp \dag    &  53.6 $\pm$ 0.38  & 49.1 $\pm$ 0.47 \\
BabbleLabel-ProgExp \dag    &  58.3 $\pm$ 1.10  & 49.7 $\pm$ 0.54 \\ \midrule
BERT-NoExp \dag       & 52.9 $\pm$ 0.97 & 49.7 $\pm$ 1.01 \\
ExpBERT \dag & 63.5 $\pm$ 1.40 & 52.4 $\pm$ 1.23 \\
\textbf{PC (BERT)} & & \\
\quad + \textbf{R} & \textbf{64.7 $\pm$ 0.62} & \textbf{53.5 $\pm$ 0.99} \\ 
\quad + \textbf{Fixed R} & not converged & 45.4 $\pm$ 1.56 \\ 
\quad + \textbf{FR} & 60.5 $\pm$ 2.59 & 49.8 $\pm$ 0.96 \\  \midrule
RoBERTa-NoExp & 62.2 $\pm$ 0.58 & 53.9 $\pm$ 0.32 \\
ExpRoBERTa & 65.8 $\pm$ 0.95 & 55.1 $\pm$ 0.31 \\
\textbf{PC (RoBERTa)} & & \\
\quad + \textbf{R}  & \textbf{66.2 $\pm$ 2.18} & \textbf{55.7 $\pm$ 0.91} \\ 
\quad + \textbf{Fixed R}  & 41.6 $\pm$ 9.31 & 50.9 $\pm$ 0.99 \\
\quad + \textbf{FR}  & 66.0 $\pm$ 2.60 & 54.5 $\pm$ 0.73 \\
\bottomrule
\end{tabular}
\caption{Results for frozen language models. The perturbed context outperforms its competitors. \dag: result from \cite{murty2020expbert}. 
}
\label{tab:freezed}
\end{table}


\subsection{Effect on Frozen Language Models}
\label{sec:exp:frozen}

For a comprehensive comparison, we also compare the effects of different models in the frozen language models. Note that the parameters of the perturbed context are fine-tunable except for the fixed random version. In addition to ExpBERT, we also compare our results with two settings of BERT + BabbleLabel as in~\cite{murty2020expbert}, which uses the outputs of the labeling functions for explanations as features (BabbleLabble-ProgExp), and the encoding of the explanations by ExpBERT as features (BabbleLabble-LangExp). We omit the results of conditional and factorized conditional perturbed contexts, as they need the trainable parameters to learn $F_i()$ and are not suitable for  frozen language models. The results are shown in Table~\ref{tab:freezed}.

The perturbed context still outperforms models with annotated explanations. This demonstrates that extra context space, rather than inductive bias, works for frozen language models. Unlike the results in Table~\ref{tab:main}, where the fine-tunable and fixed random variants are competitive, the fine-tunable random variant performs clearly better for frozen language models. We think this is because the model requires the perturbed context to provide flexibility via fine-tuning.

\subsection{Effect of the Factorization}
\label{sec:exp:factorization}

\begin{figure}[!t]
\centering
\begin{subfigure}[b]{0.225\textwidth}
	\centering
		\includegraphics[scale=.35]{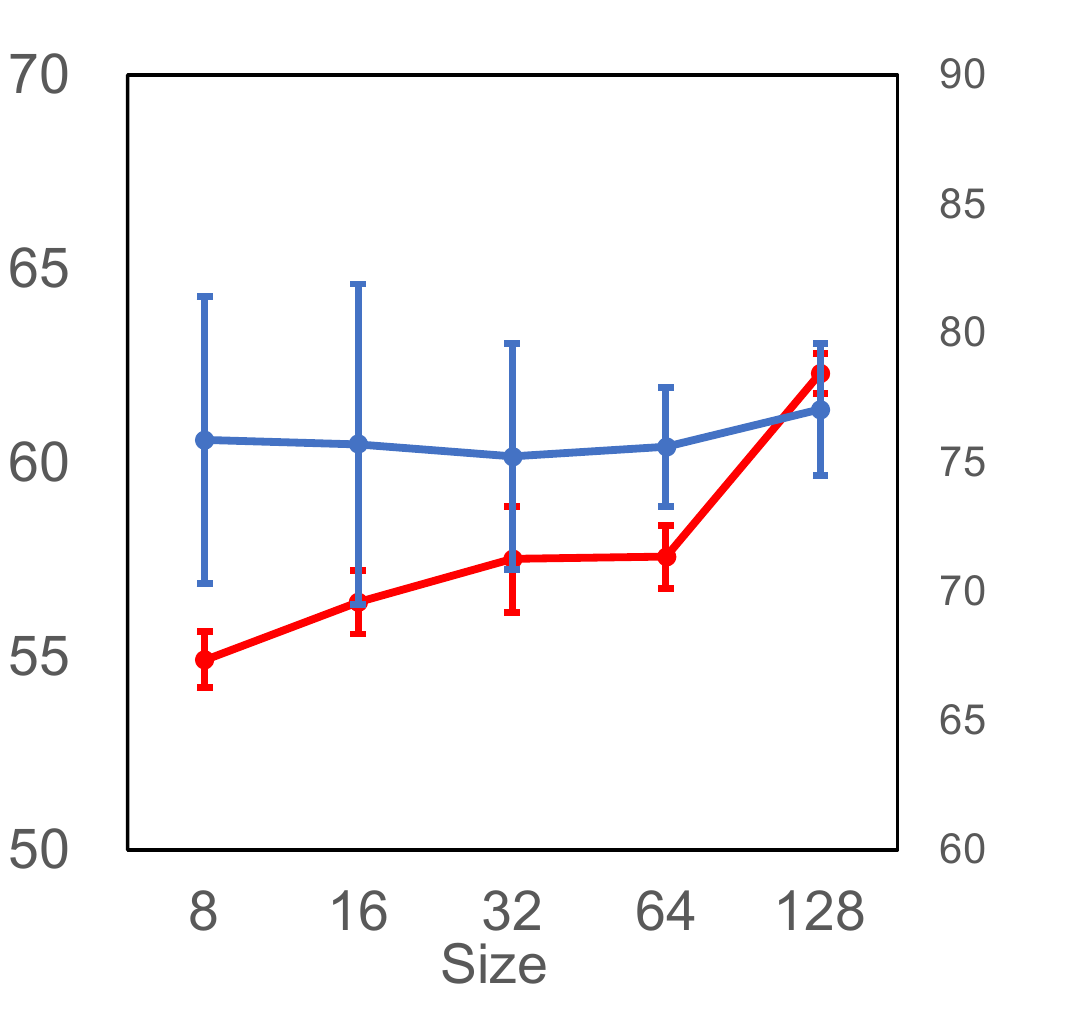}
\caption{Spouse}
\label{fig:size:spouse}
\end{subfigure}
\hspace{0.2cm}
\begin{subfigure}[b]{0.225\textwidth}
	\centering
		\includegraphics[scale=.35]{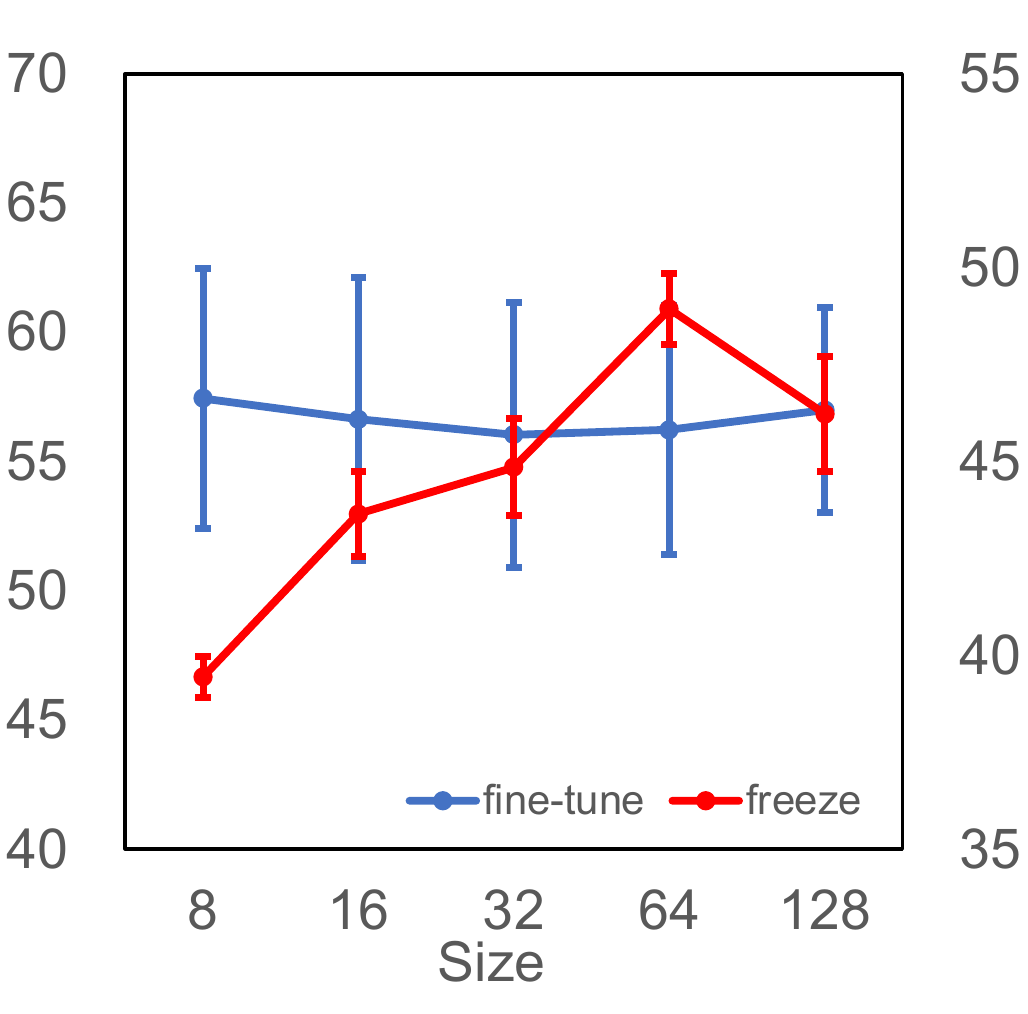}
\caption{Disease}
\label{fig:size:disease}
\end{subfigure}
	\caption{Effect of the size of the perturbed context. For frozen language models, the effect is positively correlated with size. However, this does not hold for the fine-tunable language models.}
	\label{fig:size}
\end{figure}

Results in Table~\ref{tab:main} and Table~\ref{tab:freezed} have shown that factorized models have slight performance degradation. To directly investigate the effect of flexibility/factorization, we control the size of the perturbed context of factorized models (i.e. $l$). We present the results of different $l$s in different training styles in Fig.~\ref{fig:size}.

We found that the choice of training style has a significant effect on flexibility. If fine-tuning is allowed, varying the size does not have a significant effect. However, for frozen language models, increasing the size significantly improves the results. This indicates that frozen language models require highly flexible perturbed contexts to enhance the effect. On the other hand, the tunable language model does not rely on the flexibility of the perturbed context. We think the flexibility has already been complemented by the tunable parameters in tunable language models.

\subsection{Summary} 
Considering all the results in \S\ref{sec:main_result}~\ref{sec:exp:frozen}~\ref{sec:exp:factorization}, the perturbed context has competitive or superior performance than annotated explanations if it has sufficient flexibility. The flexibility is obtained by one of the two fashions: (1) tunable perturbed context itself; (2) tunable language models.


\section{Related Work}
{\bf Introducing explanations in models} Introducing explanations in text understanding has drawn many research interests. A typical class of methods is to construct explanations for specific domains, and then convert these explanations into features and combine them in the original model~\cite{srivastava2017joint,wang2017naturalizing,hancock2018training}. For example, \citet{srivastava2017joint} use a semantic parser to transform their constructed explanations into features to apply to downstream tasks. \citet{hancock2018training} use the semantic parser as noisy labels, instead of features. \citet{murty2020expbert} argues that these semantic parsers can typically only parse low-level statements, therefore they use the language model as a ``soft'' parser to interpret language explanations, aiming to fully utilize the semantics of the explanations. 


{\bf Revisiting the value of inductive biases} Our paper presents the first proposal to revisit the inductive bias from explanations. Some previous studies worked on revisiting the inductive bias from network architectures~\cite{touvron2021resmlp,tay2021pre,liu2021pay}, and some progress has been made. For example, \citet{tolstikhin2021mlp} found that, even only using multi-layer perceptrons, competitive scores on image classification benchmarks with CNNs and Vision Transformers~\cite{dosovitskiy2020image} could be attained. \citet{guo2021beyond} found that self-attention can be replaced by two cascaded linear layers and two normalization layers based on two external, small, learnable, shared memories. \citet{melas2021you} found that applying feed-forward layers over the patch dimension obtains competitive results with the attention layer. These studies all demonstrate that it it non-trivial to integrate valid inductive biases into neural networks.
\section{Conclusion}

In this paper, we revisit the role of {\it explanations} for relation extraction. In previous studies, explanations were thought to provide effective inductive bias, thus guiding the model learn the downstream task more effectively. 
We argue that it is imprudent to simply interpret explanations' effects as inductive bias. We find that the effect of natural language explanations varies across different training styles and datasets. By randomly corrupting the explanations, we found that the effect of explanations did not change significantly as the inductive bias decreased. This suggests that the inductive bias is not the main reason for the improvement. We further propose that the key of explanation for the improvement lies in the fine-tunable context. Based on this idea, we propose perturbed contexts. Perturbed contexts do not require any annotated explanations, while still providing (fine-tunable) contexts like annotated explanations. Our experiments verified that the effectiveness of the perturbed context is comparable to that of annotated explanations, but (1) the perturbed context does not require any manual annotation, making them more adaptable; (2) the perturbed context is much more efficient than that of using annotated explanations. 
\section{Limitations}

This paper lacks a formalized analysis of the relationship among perturbed contexts/pre-training/model generalization. Although we try to analogize pre-training and prompt in Appendix~\ref{sec:method:rationale} to explain how the perturbed context works, it lacks a rigorous mathematical description.

The validation of the perturbed context is limited to relation extraction. Although we show its potential on other applications in Appendix~\ref{sec:exp:beyong}, the experiments are still primitive. A more systematic evaluation on different NLP tasks is still excepted.

\bibliography{sample-base}
\bibliographystyle{ACM-Reference-Format}

\appendix

\section{Hyperparameters} 
\label{sec:hyper}
We use the {\tt bert-base-uncased} and {\tt roberta-base} from Huggingface transformers~\cite{wolf2020transformers}. We set the batch size $=32$, learning rate $=2e-5$, and train the model for 5 epochs for all three relation extraction tasks. By default, we set the size of the intermediate space of the factorized models to $l=32$. 8 NVIDIA RTX 3090Ti GPUs are used to train the models.

{\bf Initialization} For randomly perturbed context, we empirically found that the initialization of $M$ will affect the results. After some trials, we found that initializing these parameters using a normal distribution with the mean and variance as in the token embeddings of the vanilla BERT is a practical choice.

\section{Rationale and Relationship to Prompt Tuning}
\label{sec:method:rationale}
Our proposed perturbed contexts can be considered as fine-tunable contexts that guide model training. Prompt-tuning is a similar approach using fine-tunable languages. We compare their differences here.

{\bf Prompt tuning}~\cite{wei2021finetuned,schick2021s,shin2020eliciting,li2021prefix,brown2020language} utilize the pre-trained masked language modeling task and map the predictions of $\mathrm{[mask]}$ to the target label. For example, predicting {\tt good} for the mask in {\tt I love this movie. Overall, this is a $\mathrm{[mask]}$ movie} will classify it into a positive sentiment. However, for the relation extraction task of interest in this paper, it is difficult to establish the mapping between $\mathrm{[mask]}$ and relations by prompt due to the large label space~\cite{chen2021adaprompt}. Our approach, on the other hand, can be applied to arbitrarily complex sentence classification tasks since the sentence representation is obtained directly from the $\mathrm{[CLS]}$ token. 


{\bf Rationale} The rationale for prompt tuning is that the pre-trained masked language model has a strong generalization ability for $\mathrm{[mask]}$ prediction. Therefore, the prompt performs well in the few-shot setting. Our perturbed context, on the other hand, exploits the generalization ability of the pre-trained language model for {\it contextual representation}. That is, given the target $sentence + context$, the language model can efficiently use a richer context to augment the target sentence. Based on the generalization ability for the rich context, given a fine-tunable perturbed context, the language model can automatically learn the optimal perturbed context as the context.

\section{Effect of Multiple Perturbed Contexts} 
\label{sec:exp:multi}

Although we mainly discuss the scenario of a single perturbed context above, previous work~\cite{murty2020expbert,hancock2018training} have used multiple explanations. Therefore, we also validate the effect of using multiple perturbed contexts. Specifically, we formulate $n=5$ randomly perturbed contexts:
\begin{equation}
    e_{vi} \coloneqq \mathrm{o_1\:\: [V]^i_{1}\:\: [V]^i_{2}\:\: \cdots\:\: [V]^i_{m} \:\: o_2} \text{ for } i=1 \cdots n
\end{equation}
Then, we append each perturbed context $e_{vi}$ to the original sentence $s$ and represent these $n$ sentence pairs as in ExpBERT. We concatenate the representations of all sentence pairs to form the resulting feature vector, and use an MLP over it to conduct the classification.

The results of multiple randomly perturbed contexts are shown in Table~\ref{tab:multiple}. We found that the improvement using multiple randomly perturbed contexts is not significant. We consider that this is because a single perturbed context already provides enough fine-tunable context.

\begin{table}[!tb]
\small
\centering
\begin{tabular}{lccc} \toprule
 & Spouse & Disease  & TACRED  \\ \hline
Single    & \textbf{76.7} & 57.3 & \textbf{67.6} \\
Multiple     & 75.1 & \textbf{57.4} & 67.4 \\ \bottomrule
\end{tabular}
\caption{Comparison results of single/multiple perturbed contexts. Results are averaged over 5 runs. Using multiple perturbed contexts does not show surpassing effects.}
\label{tab:multiple}
\end{table}

\section{Perturbed Contexts as Augmented Context? Application beyond Relation Extraction}
\label{sec:exp:beyong}

Notice that our proposed perturbed contexts are actually fine-tunable contexts added to the original sample, which does not correspond to the semantics of any actual explanation. Therefore, it is natural to think that these perturbed contexts can be used not only as an alternative to annotated explanations for relation extraction, but also for broader applications.

It may be obvious that adding external  relevant context can improve the representation of the target text. This idea has been verified on several tasks such as reading comprehension~\cite{long2017world}, entity linking~\cite{logeswaran2019zero}, and even image classification~\cite{radford2021learning}. One typical class of the external context is the knowledge description of entities. The model will jointly represent the target text and the descriptions of the entities within it to enhance the text representation.

In this section, we made a preliminary attempt at two fundamental tasks that involve entity knowledge: open entity typing (OpenEntity~\cite{choi2018ultra}) and word sense disambiguation (WSD~\cite{raganato2017word}). We study the effect of replacing knowledge-related contexts with perturbed contexts.

\subsection{Tasks}
\textbf{Entity typing} Given a sentence $s$, the goal of entity typing is to classify an entity $ent$ in $s$. For example, for the sentence \textit{Paris is the capital of France.} and the target entity {\it Paris}, the model is required to classify {\it Paris} into \textit{Location}. We propose to use perturbed context as text augmentation for entity typing. To address this, we construct the randomly perturbed context in the form of $e_{v} \coloneqq \mathrm{[V]_{1}\:\: \cdots\:\: [V]_{m} \:\: ent \:\: [V]_{m + 1}\:\:  \cdots\:\: [V]_{2m}}$. Then, we refer to Eqn.\eqref{eqn:input} to classify the augmented text by BERT.

\textbf{WSD}  Given a sentence $s=w_1, \cdots ,w_n$, and a polysemy word $w_t (1 \le t \le n)$ with candidate senses $\{c_1, c_2, ..., c_m\}$, WSD aims to find the sense for $w_t$. For instance, for the sentence \textit{Apple is a technology company.} and the target polysemy word \textit{Apple}, the model needs to recognize whether it refers to a fruit or a technology company. To augment the model effectiveness on WSD, we construct the perturbed context and model similar to the entity typing task. The only difference is, we follow~\cite{huang2019glossbert} to use the final hidden state of the target word to conduct the classification.

\subsection{Setup}

{\bf Baselines} For entity typing, We consider two types of baselines: the first type directly fine-tunes the target task. These baselines include BERT and NFGEC~\cite{shimaoka2016attentive}. The second type first train the model over the joint corpus of the text and the external knowledge, and then fine-tune it on the target task. These baselines include  KnowBERT~\cite{peters2019knowledge} and ERNIE~\cite{zhang2019ernie}. KnowBERT enhances contextualized word representations with attention-based knowledge integration using WordNet~\cite{miller1995wordnet} and Wikipedia. ERNIE integrates knowledge through aligning entities within sentences with corresponding facts in Wikidata~\cite{vrandevcic2014wikidata}. For WSD, we consider vanilla BERT as our baseline.

{\bf Hyperparameters} We choose the same hyperparameters as in the relation extraction tasks, except that we train our model for 10 epochs for OpenEntity and 6 epochs for WSD, respectively. For WSD, we refer to the previous setting of training/valid/test splits~\cite{zhong2010makes, iacobacci2016embeddings}.

\subsection{Results}

\begin{table} [!tb]
\small
\centering
\setlength{\tabcolsep}{1.4mm}{
\begin{tabular}{l|c|cccc|cccc|c}
    \toprule
    & \multicolumn{1}{c|}{Dev} &
    \multicolumn{4}{c|}{Test Datasets}\\
     & SE07 & SE2 & SE3 & SE13 & SE15 & \bf All \\
    \hline
    BERT\dag & 61.1 & 69.7 & 69.4 & 65.8 & 69.5 & 68.6 \\
    Our BERT & 64.8 & 73.1 & 71.7 & 68.2 & 73.6 & 71.2 \\
    \textbf{PC (BERT) + R} & \textbf{66.2} & \textbf{74.6} & \textbf{72.5} & \textbf{68.4} & \textbf{74.0} & \textbf{72.0} \\
    \bottomrule
    \end{tabular}
    }
    \caption{Results on WSD datasets. \dag: results from \cite{huang2019glossbert}.} 
    \label{tab:wsd}
\end{table}

\begin{table}[!t]
\small
\centering
\resizebox{\columnwidth}{!}{
\begin{tabular}{l|c|ccc}
\toprule
Model & Joint pre-train & \textbf{P} & \textbf{R} & \textbf{$\textbf{F}_1$} \\ \midrule
BERT \ddag & No & 76.37 & 70.96 & 73.56 \\
Our BERT  & No & 75.98 & 73.42 & 74.68 \\
NFGEC \ddag & No & 68.80 & 53.30 & 60.10 \\
ERNIE \ddag  & Yes & 78.42 & 72.90 & 75.56 \\
KnowBERT \ddag & Yes & \textbf{78.60} & \textbf{73.70} & \textbf{76.10} \\ \midrule
\textbf{PC (BERT) + R} & No & 77.42 & 72.95 & 75.12 \\ \bottomrule
\end{tabular}}
\caption{Results on OpenEntity. \ddag: results from \cite{zhang2019ernie}.}
\label{tab:entity_typing}
\end{table}

The results of the randomly perturbed context on WSD and OpenEntity are shown in Table~\ref{tab:wsd} and Table~\ref{tab:entity_typing}, respectively. Our proposed approach still outperforms the baselines without joint pre-training. Even compared with baselines that use joint pre-training for knowledge integration, the performance degradation is not significant. This indicates that, to some extent, the perturbed context enhances the representation of texts that require entity knowledge.
This shows the potential of our approach in different scenarios. We leave it to future work to explore the effects of the perturbed context on more tasks.

\end{document}